\documentclass[runningheads]{llncs}
\usepackage[super]{nth}
\usepackage{placeins}
\usepackage{float}
\usepackage[T1]{fontenc}
\usepackage[english]{babel}
\usepackage{graphicx}

\usepackage{algpseudocode}
\usepackage{algorithm}

\usepackage{hyperref}
\usepackage[sorting=none]{biblatex}
\usepackage{csquotes}
\usepackage{amsmath}

\usepackage{caption} 
\algnewcommand\algorithmicforeach{\textbf{for each}}
\algdef{S}[FOR]{ForEach}[1]{\algorithmicforeach\ #1\ \algorithmicdo}
\algdef{SE}[REPEATN]{RepeatN}{End}[1]{\algorithmicrepeat\ #1 \textbf{times}}{\algorithmicend}

\AtEveryBibitem{
    \clearfield{urlyear}
    \clearfield{urlmonth}
}

\authorrunning{Deproost et al.}% Part of LEFT running header
\titlerunning{Critic-Moderated Genetic Programming}% Part of RIGHT running header

\begin{document}
\title{Human-Readable Programs as Actors of Reinforcement Learning Agents Using Critic-Moderated Evolution}

\author{ Senne Deproost\inst{1}\orcidID{0009-0009-4757-0290}\and \\
 Denis Steckelmacher \inst{1}\orcidID{0000-0003-1521-8494}\and
Ann Now\'{e}\inst{1}\orcidID{0000-0001-6346-4564}}

\institute{Vrije Universiteit Brussel, Pleinlaan 2, 1060 Elsene, Belgium}
\maketitle              % typeset the header of the contribution

\begin{abstract}
With Deep Reinforcement Learning (DRL) being increasingly considered for the control of real-world systems, the lack of transparency of the neural network at the core of RL becomes a concern.
Programmatic Reinforcement Learning (PRL) is able to to create representations of this black-box in the form of source code, not only increasing the explainability of the controller but also allowing for user adaptations.
However, these methods focus on distilling a black-box policy into a program and do so after learning using the Mean Squared Error between produced and wanted behaviour, discarding other elements of the RL algorithm. The distilled policy may therefore perform significantly worse than the black-box learned policy.
In this paper, we propose to directly learn a program as the policy of an RL agent. We build on TD3 and use its critics as the basis of the objective function of a genetic algorithm that syntheses the program. Our approach builds the program during training, as opposed to after the fact. This steers the program to actual high rewards, instead of a simple Mean Squared Error. Also, our approach leverages the TD3 critics to achieve high sample-efficiency, as opposed to pure genetic methods that rely on Monte-Carlo evaluations.
Our experiments demonstrate the validity, explainability and sample-efficiency of our approach in a simple gridworld environment.

\end{abstract}

\section{Introduction}
% Motivation for XRL
While Deep Reinforcement Learning (DRL) becomes a viable method to generate system controllers in an automatic manner \cite{bellemare_etal_2020_AutonomousNavigation, degrave_etal_2022_MagneticControl}, their broader adoption becomes hindered by the lack of transparency and explainability. Since the DRL agent behaviour is computed by a black box model, such as a neural network, the exact method used by the network to map a state to an action is often too complex and beyond human comprehension \cite{rudin_2019_StopExplaining, miller_2019_ExplanationArtificial}. For control engineers who require specific guarantees from the controller  (stability, robustness, ...) this lack of transparency is unacceptable, leading to low adoption of DRL in their workflow \cite{osinenko_etal_2022_ReinforcementLearning}. \\
% Moving from rulesets to programs 
A decade after the first successes of merging deep and reinforcement learning \cite{mnih_etal_2015_HumanlevelControl}, the emerging field of Explainable Reinforcement Learning (XRL) introduced both local and global explanations within the different stages of training an agent \cite{bekkemoen_2023_ExplainableReinforcement}. Whereas \textit{global explanation} methods consider the policy as a whole, and describe its behavior in every state at once (such as "avoid obstacles by the North"), \textit{local explanations} focus on a particular time-step, and offer explanations such as "I went up because there is an obstacle in front". To be able to generate good controllers, a human-readable \textit{global} mapping from input to output is needed \cite{langer_etal_2021_WhatWe}. To represent this, different types of explanations have been considered, for instance rule-based \cite{engelhardt_etal_2023_SamplebasedRule}. These rules are queried as part of a rule set or selected in a hierarchical way using decision trees \cite{Coppens2019DistillingDR}. Recently, program-like representations have been considered, closely resembling regular computer programs \cite{verma_etal_2018_ProgrammaticallyInterpretable, trivedi_etal_2022_LearningSynthesize, hein_etal_2018_InterpretablePolicies}. They contain the same building blocks as regular source code (variables, control flow, operators, ...) with statements being executed in a sequential order. This improves on both readability of the explanation as well as adaptability by the user. Furthermore, if the program is expressed in a syntax compatible with a Programmable Logic Controller (PLC), their deployment onto actual hardware becomes straightforward \cite{guo_etal_2017_SymbolicExecution}.

% Genetic programming
The realization of programs within this Programmatic Reinforcement Learning (PRL) approach is still challenging. Current state-of-the-art methods such as generative networks \cite{trivedi_etal_2022_LearningSynthesize} and search-based templates \cite{verma_etal_2018_ProgrammaticallyInterpretable} have shortcomings in their ability to expand beyond a fixed set of possible solutions. To relax this restriction, another approach is to take inspiration from the field of program synthesis using Genetic Programming \cite{hein_etal_2018_InterpretablePolicies}, motivated by the fact that producing a program is an optimization problem without gradients available, and Genetic Programming is both easy to apply and works well for those problems.

% Our approach
\textbf{Contribution: }In this paper, we improve on previous approaches closely related to model distillation \cite{hinton_etal_2015_DistillingKnowledge} by proposing a reward-driven optimization. Instead of minimizing the error between the prediction error of the original model and the distilled one, we opt for the exploitation of the critic network in a TD3 agent \cite{fujimoto_etal_2018_AddressingFunction} to compute the gradient of the program quality (average Q-Values it encounters) with regards to actions predicted by the program. We then steer the genetic algorithm to move the actions produced by the program in the direction of that gradient.

We observe that it is possible to generate programs for simple a simple gridworld environment, with highly promising sample-efficiency, policy quality and explainability.

\section{Background}
\subsection{Reinforcement Learning}
Reinforcement Learning (RL) is a machine learning paradigm used to solve sequential decision problems where each step is taken at a timestep $t \in \left[0, t_{max} \right]$ \cite{sutton_barto_2014_ReinforcementLearning} wit $t_{max}$ as time horizon. These problems are modeled as a Markov Decision Process (MDP) \cite{bellman_1957_MarkovianDecision}, a control problem scheme represented by a tuple $ \left ( S, A, P_a, R_a \right )$ with $S$ the state space, $A$ the action space, $P_a$ the probability distribution of state transitions given action $a$ and $R_a$ the immediate reward given after performing $a$ in the environment. The transition between states is described by $\mathcal{P}^a_{ss'}$. For the formulation of an MDP, we assume the transitions of states to be Markovian, with $Prob\left \{s_{t+1} = s',   r_{t+1} = r  \mid s_t, a_t \right \}$. This means the transitions of the states are not influenced by past transitions.

At any timestep in the MDP a state $s_{t}$ is observed. The policy  $\pi_t$ predicts an action $a_t$ for that observation. A new state of the environment $s_{t + 1} \in S $ is observed together with reward signal $r_{t + 1} \in R $. This can consist of both positive and negative values and is provided by the user as reward function $\mathcal{R}^a_{ss'}$ expressing the control objective. The objective of a Reinforcement Learning agent is to learn a policy that maximizes the discounted sum of rewards $R(\tau) = \sum\limits_{t} \gamma^t r_{t + 1}$, with $\gamma \in \left[0, 1\right[$ as the discount factor. 

\subsection{Q-learning}
A well-known approach to Reinforcement Learning, at the basis of our work, is Q-Learning. The quality of an action $a_t$ under a policy $\pi$ is given by a Q-Value $Q_{\pi}(s, a)$, the expected return obtained by executing the $a_t$ in $s_t$, and then following policy $\pi$. The Reinforcement Learning agent iteratively refines estimates of the Q-Values. The update rule is given by:

\begin{align*}
    \delta &= r_{t+1} + \gamma Q(s_{t+1}, a_{t+1}) - Q(s_t, a_t) & \text{(TD error)} \\
    Q(s_t, a_t) &\gets Q(s_t, a_t) + \alpha \delta
\end{align*}

\noindent
with learning rate $\alpha \in \left ] 0, 1 \right ]$. Selecting an action $a_t$ under the optimal policy $\pi^*(s_t)$ is obtained by $a_t = \operatorname{argmax}_a Q^*(s_t, a)$ with $Q^*$ the converged optimal Q-Value function.

\noindent

\subsection{Policy gradient}

Q-Learning is a value-based RL method because it learns the value of actions (how good they are). Another approach to RL is Policy Gradient, a policy-based RL method, that directly learns the best parameters of a parametric policy $\pi_{\theta}$, such that the agent achieves the highest-possible returns \cite{sutton1999policy}:

\begin{align*}
    R_t &= \sum_{t'=t}^{T_{max}} r_{t'} \\
    \nabla_{\theta} &= \sum_t R_t \log \pi(a_t | s_t)
\end{align*}

\noindent
with $T_{max}$ the time-step at which one episode finishes. Policy gradient using $\nabla_{\theta}$ is applied after the episode, as it needs the Monte-Carlo sum of rewards obtained during the episode. After each application of Policy Gradient, any samples of states, actions and rewards needs to be discarded and new ones need to be collected by executing the updated policy in the environment. This causes Policy Gradient to have low sample-efficiency.

\begin{figure}[t]
    \centering
    \includegraphics[width=\linewidth]{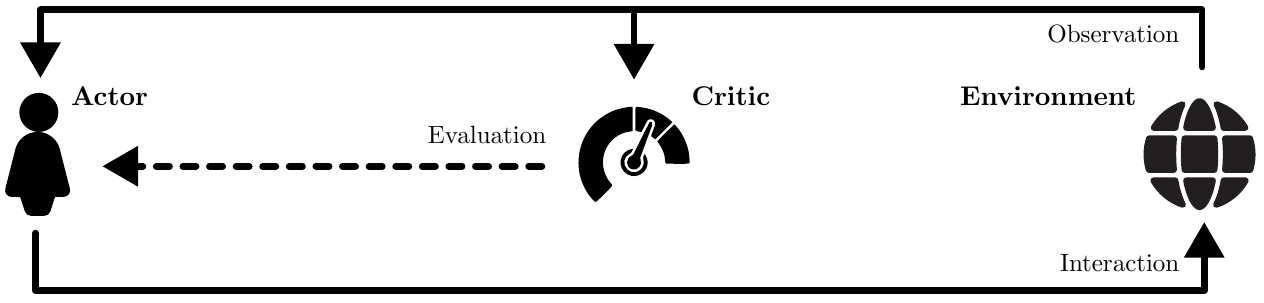}
    \caption{Schematic overview Actor-Critic architecture. The critic provides Q-values to the made interactions while the actor updates the policy via policy gradient \cite{fujimoto_etal_2018_AddressingFunction}.}
    \label{fig:TD3}
\end{figure}

Recent advances in policy-based methods led to Actor-Critic methods that combine an explicit policy $\pi_{\theta}$ with Q-Values (the critic). An example of Actor-Critic algorithm is TD3 \cite{fujimoto2018addressing}, graphically summarized in Figure \ref{fig:TD3}. TD3 trains two separate critic networks with SARSA, and then queries those for a gradient of the actions in a batch of states: $\nabla_a Q(s, \cdot)$. This gradient is then back-propagated into the actor so its parameters are changed in a way that produce actions that are evaluated as more promising by the critics. The use of multiple critics rather than one is one of the improvements of TD3 on Deep Deterministic Policy Gradient (DDPG) \cite{Lillicrap2015ContinuousCW}, reducing value overestimation.

%\begin{align*}
%    \nabla_{\theta} &= Q(s, a = \pi(s))
%\end{align*}

\subsection{Use of black-box models in RL}
% Why deep?
Traditionally, an RL policy is represented by a table of Q-values for discrete state and action spaces. To learn in continuous spaces, a discretization function needs to be applied. The resulting discretization error could be minimized by extending the table size at the cost of memory, which explodes for multi-variable observation spaces. Replacing the table with an artificial neural network addresses this issue since it is capable to generalize between encountered inputs and have been considered universal function approximators\cite{hornik_etal_1989_MultilayerFeedforward}. 

% Disadvantage
However, due to their outputs being the result of a large number of simple operations (additions, multiplications, simple non-linear functions), using neural networks comes at the cost of losing model transparency, due to the sheer number of operations and seemingly-arbitrary numbers appearing in neural networks. To address this lack of \textit{explainability}, the recent field of Explainable Reinforcement Learning (XRL) has prompted many researchers to come up novel methods of representing policies or critics \cite{bekkemoen_2023_ExplainableReinforcement}. Various algorithm generate explanations at different moments (from the beginning, during or after training), and consider either global or local explanations. In this paper, we focus on global explanations, with the aim of producing a program that can be "copy-pasted" in a Programmable Logic Controller, read, understood and trusted by the control engineer.

%\subsection{Genetic Programming}
% Bio-mimick
%Genetic Programming is a gradient-free function optimization method. Several optimization techniques are the result of nature-inspired phenomena. Often these biomimicking methods rely on complex emergent behaviour from the interaction of many simpler processes. Noticeable examples are Ant Colony Optimization (ACO) \cite{dorigo_stutzle_2004_AntColony} and Swarm Intelligence \cite{kennedy_eberhart_2001_SwarmIntelligence}. The first being inspired by the coordination of ant societies while the latter is a broader group of approaches where huge groups of simple simple individualistic behaviours are contributing to solve a complex problem. Reinforcement Learning on itself has roots in nature. It uses reward in an iterative manner, as demonstrated by Thorndike \cite{thorndike_1898_AnimalIntelligence}, and also mimics the brain with the neural network model in DRL. 

%Another bio-inspired function optimization technique is Genetic Programming (GP) \cite{koza_1992_GeneticProgramming}. These are based on Darwinian evolution among living species, guided by the principles of the survival of the fittest. Aside using a large amount of individuals, this approach requires many iterations called \textit{generations} that iteratively creates, changes and destroys candidate solutions in a population, using concepts borrowed from genetics, such as DNA, mutations, crossover and parents producing offsprings.

\subsection{Genetic Programming}
% How evolution starts
Optimizing via evolution is performed by executing iterations of the Genetic Programming (GP) process or generation loop. The starting condition is a population of individuals with $g_i$ being the $i$th individual's genome. This genome represents model parameters to be optimized (polynomials, weights, ...) or an encoding of the model itself (program statements). \textit{It is up to the user to translate the problem into a representation that fits the GP process}, and one contribution of our paper is the representation of programs as a sequence of real-valued genes. A fitness function is defined, encapsulating the objective to be optimized in order to solve the problem.

% How evolution continues
At the beginning of each generation, a number of individuals are selected in order to perform \textit{crossover}. These \textit{parents} will exchange genetic material with each other to produce offspring that share their characteristics. Afterwards, \textit{mutation} is applied to the whole population to encourage random exploration in the genome. From this phase, advantageous behaviour can arise that is passed through the generations. It is often used to look beyond the peaks of local optima. In the \textit{selection} phase, the fitness function is applied on each individual. Those that perform worst are eliminated. The best scoring individuals survive and will contribute to the next generational loop.

% How it ends
The optimization ends when the user-specified number of generations is met or some performance threshold. The individual achieving the highest fitness is considered as the best individual. However, in the context of generating programs, individuals with similar performances could also be considered if other criteria are considered (program length, limited nesting, ...). %Having shorter programs with the same performance as longer ones is of great interest when explainability towards a user is considered.

\section{Related work}
% - Work done on programmatic RL
% - Work done on genetic RL
%      - Creating models
%      - Finding parameters for neural networl
% - Work done at intersection (hein)

\subsection{Programmatic Reinforcement Learning}
In the recent past, several attempts have been made to synthesize programs from an RL agent.
% Verma
Verma et al. introduced a template-based search over a set of programmatic policies \cite{verma_etal_2018_ProgrammaticallyInterpretable}. By using an \textit{oracle} network from a trained DRL agent, they steered the search for fitting variable values. However, this method is not so flexible as the templates are user-defined at the start of training, which could be non-optimal. Nevertheless, they showed the approach is well suited for PID-like programs on both a racing game and a classic control environments.
% Trivedi
Trivedi et al. propose to learn program embeddings using a variational autoencoder (VAE) \cite{trivedi_etal_2022_LearningSynthesize}. This model first learns program embedding by reconstructing source code using several loss functions. Afterwards, it can be used to directly learn a programmatic policy by interacting in the environment and suggesting candidate programs based on return maximization.
% Hein
Hein et al. incorporate genetic programming to generate programmatic trees that represent simple algebraic equations \cite{hein_etal_2018_InterpretablePolicies}. Mutation of genes happen in the nodes of the trees while crossover switches position between two subtrees within the parents, ensuring the program is kept valid. This method was applied on both the CartPole and MountainCar environment with well performing programs emerging at certain complexity levels. \\

\subsection{Genetic Reinforcement Learning}
% Reward
Genetic programming, besides evolving surrogate models, has been used to optimize other components of the Reinforcement Learning process. Niekum et al. created new reward functions using PushGP, a stack-based programming language \cite{niekum_etal_2010_GeneticProgramming}. Using a set of simple operators, they could formulate a better reward for dynamic gridworld environments. This approach captures common features among the different environments, allowing for hierarchical decomposition.
% Hyperparameters
Finally, Sehgal et al. used GP to optimize the hyperparameters of a DDPG agent on MuJoCo robotic environments \cite{sehgal_etal_2019_DeepReinforcement, todorov_etal_2012_MuJoCoPhysics}. \\

\section{Evolving programs}
Our contribution builds on TD3 and Genetic Programming to implement the actor of a Reinforcement Learning agent as a program, expressible in source code. Our method differs from related work in two key aspects:

\begin{enumerate}
    \item The program replaces the TD3 actor, and is optimized using gradients of actions produced by the critics. This allows the actor to directly optimize the returns obtained by the agent, as opposed to approximate some black-box policy, leading to high-quality programs being produced in terms of reward.
    \item The program is trained using gradients produced by the TD3 critics, not by direct interaction with the environment. This makes our method several orders of magnitude more sample-efficient than other programmatic RL approaches, that evaluate individuals in the Genetic Algorithm population by performing rollouts in the environment.
\end{enumerate}

We now introduce how we represent programs as a list of real values (the genome), execute these programs, and integrate them as the actor of TD3.

\subsection{Representing programs as sequences of real values}
\begin{figure}
    \centering
    \includegraphics[width=\linewidth]{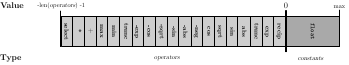}
    \caption{Values in the gene space and their encoding. Positive values represent literals while negative values the different operators. Literals have a sign randomly sampled at run time. The sign is fixed with $abs$ or $-abs$.}
    \label{fig:genome_space}
\end{figure}

% Encoding
At the start of the training process, a population of genomes is initialized as a two-dimensional array of size $\texttt{num\_individuals} \times \texttt{num\_genes}$. Each $i$th genome $g_i \in G_{min}^{max}$ is a selection of \texttt{num\_genes} random floats with values in the range of $\left]-\texttt{len(operators)} , \texttt{max} \right]$  from the gene space $G$ which represents an encoding of an arbitrary program. As shown in figure \ref{fig:genome_space}, we opted for a simple set of operators, encoded as negative values, and literals of randomly-sampled sign, encoded as positive values.

The sequence of operators and literals that make a program are interpreted as the postfix notation of a program, directly linked to the evaluation of the program by a stack machine. For instance, "3 2 sin +" represents $sin(2) + 3$ in the usual infix notation: it pushes 3, pushes 2, replaces the 2 with $\sin(2)$, then pops $\sin(2)$ and 3 and replaces it with $\sin(2) + 3$.

Some operators have a limited domain, such as the square root operator. We want to ensure that every program is runnable, and thus implement some default value for the operators that have inputs outside of their domain. \texttt{reciprocal}, or $\frac{1}{x}$, resolves to the value 20 when $x$ is between -0.05 and 0.05. \texttt{exp} clamps its input to at most 10. For the square root, the return value is 0 should the operand value be negative.

\subsection{Optimization landscape}

The representation of the programs has been designed with some specific aspects that improve learning:

\begin{enumerate}
    \item Abrupt changes in program behavior following a small mutation of a gene makes the optimization landscape rigid, with many plateaus. To address this issue, we add some stochastic aspects in the program \ref{eq:cast}.
    \item To prevent random programs (when the agent has not learned yet) from biasing the behavior of the agent, we ensure that random programs produce an expected value of 0. We do that by ensuring that literals have no definite sign (the sign is sampled at random at runtime), so they are not biased towards positive values. Functions that have an image biased towards positive or negative numbers also have a negated version. For instance, there exist sqrt and -sqrt.
\end{enumerate}

To smooth the optimization landscape, we introduce stochasticity in the mapping from real value to operators. Instead of casting the real value to an integer, and using that integer to identify an operator, we identify the operator with:

%\begin{align*}
\begin{equation}
        o = \left \lfloor g + x \sim \mathcal{U}(-0.5, 0.5) \right \rfloor
\label{eq:cast}
\end{equation}
%\end{align*}

\noindent
with $o$ the integer index of an operator, $g$ the corresponding real-valued gene, and $x$ a value uniformly sampled from the -0.5 to 0.5 range. When programs are evaluated, they are actually sampled and run 10 times, and the output of the 10 runs is averaged. Slowly changing the value of genes now has the effect of slowly changing that average value, leading to a smooth optimization landscape.

\subsection{Execution}
% Stack based execution and end operator
To execute a program, a simple stack-based execution is used. Given is the input observed state $s$, and a genome $g_i$. The stack is pre-populated with $[s] \times 20$, so input values are available many times on the stack. This encourages operators at the beginning of the genome to use state variables, leading to more state-dependent and \textit{reactive} programs. Genes are then interpreted as described in the previous section: either as literals of random sign, or as operators sampled as described above.

Literals are pushed onto the stack. When an operator is encountered, as many values as required operands are popped from the stack. If a stack underflow happens, the program is considered invalid and given a low fitness. When the proper amount of operands is retrieved, the operator is applied and the result pushed back on the stack.

At the end of execution, the result of the program is the current top of stack value.

\subsection{Critic-Moderated evolution}
\begin{figure}[t]
    \centering
    \includegraphics[width=\linewidth]{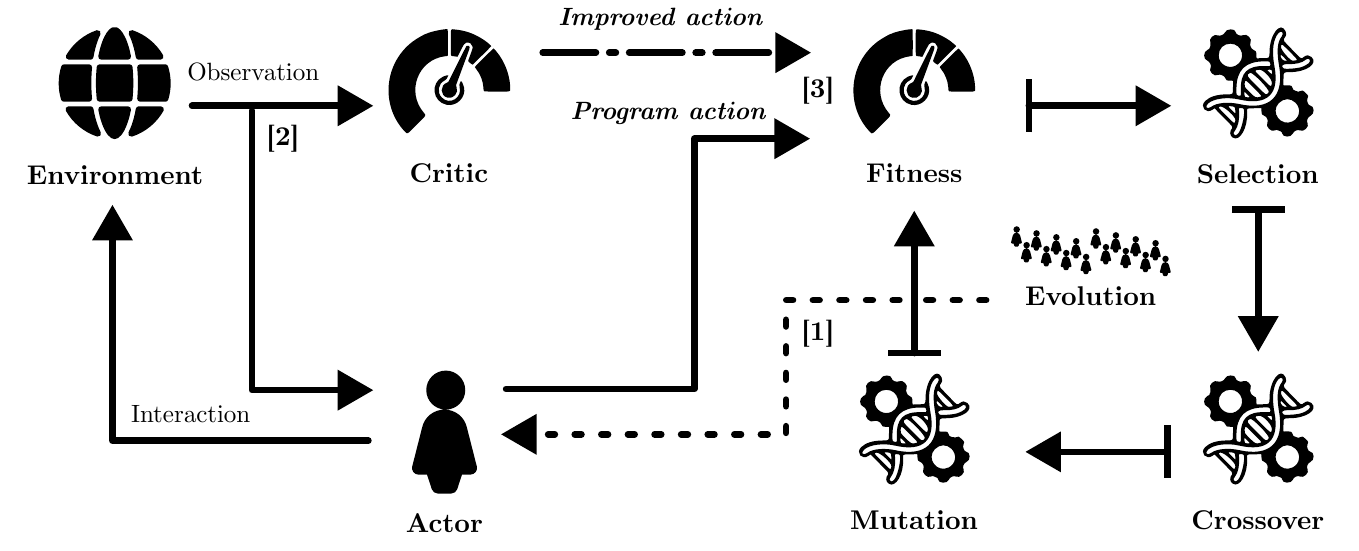}
    \caption{Our extension of TD3 with the evolution-based optimizer. With critic-improved actions, the evolutionary loop is performed to optimize the population of candidates.}
    \label{fig:TD3_extended}
\end{figure}

% PyGad
We provide an overview of our method in figure \ref{fig:TD3_extended} together with pseudocode \ref{alg:cm-gp}.
To perform the evolutionary loop, we use the PyGad library \cite{gad_2023_PygadIntuitive}. Used hyperparameters during the evolutionary loop can be found in appendix \ref{tab:hparams}.

In this paper, we assume both the state space and action spaces to be continuous. When the action space has several dimensions, we learn one program per action dimension.

Every \texttt{policy\_freq} time-steps, we optimize the programs according to the following procedure. We first use both critics of TD3 to produce \textit{improved actions} $A^*$ for a batch of states $S$. For this, we query the current programs for current actions $\hat{A}$ for $S$. We then ask the two critics for Q-Value estimates $Q_A(S)$ and $Q_B(S)$. We compute an overall \textit{program quality} metric by averaging the Q-Values over $Q_A$ and $Q_B$, and over states. This produces a single real value. All these operations are performed with autograd enabled (using PyTorch in our case \cite{paszke_etal_2019_PyTorchImperative}) which automatically computes gradients when performing backward passes through the critics. We can then retrieve the gradient of that real value with regards to the actions $\hat{A}$ produced by the programs, leading to $\nabla_A$. Slightly improved actions $A^*$ are then produced by computing $A^* = \hat{A} + \nabla_A$.

\begin{algorithm}[H]
\caption{Critic-Moderated Genetic Programming (CM-GP)}
\begin{algorithmic}[1]

\Require TD3 critics $Q_A$ and $Q_B$, population $g$, rollouts \texttt{buffer}, environment \texttt{env}
\For{$n = 0$ to \texttt{n\_steps}}
    \State $g^*$ = \texttt{best\_individual($g$)}                            \Comment{[1] Realize best program}
    \State $p_{g^*} \gets g^*$ 
    \State $\texttt{buffer} \gets \texttt{collect\_rollout(env)}$ using $p_{g^*}$
    \State Sample $rollout$ from \texttt{buffer}
    \State \texttt{train($Q_A, Q_B$)} with $rollout$
    \If{$step \mod \texttt{policy\_freq} = 0$}
        \State $S \gets \operatorname{states}(rollout)$
        \State $\hat{A} \gets p_{g^*}(S)$                                        \Comment{Actions of the current best program}
        \RepeatN{$50$}                                                           \Comment{[2] Compute improved actions}
            \State $\nabla_A = \operatorname{mean}(Q_A(S), Q_B(S))$ 
            \State $A^* = \hat{A} + \nabla_A$
            \State $\hat{A} \gets A^*$
        \End
        \State \texttt{optimize($g, \hat{A}, S$)}                                \Comment{[3] Generation of evolution}
    \EndIf
\EndFor

\end{algorithmic}
\label{alg:cm-gp}
%\caption{Pseudocode of the algorithm with references to figure \ref{fig:TD3_extended}.}
\end{algorithm}

We now set $\hat{A} \gets A^*$ and repeat the process above 50 times. This forms a sort of 50-step gradient ascent algorithm that, starting from the current output of the programs, follows the TD3 critics to lead to better actions. This optimization process on a critic is akin to the CACLA method proposed in \cite{van2007reinforcement}. We stop that process if $A^*$ becomes too different from the actions produced by the programs (when the L1 norm is above 1), akin to the trust region of TRPO proposed in \cite{schulman2015trust}.

The improved actions $A^*$ can now be used to optimize the programs with the Genetic Algorithm. The fitness function on a batch of states $S$ for an individual $g_i$ in the population is the mean-squared error (MSE) between program actions $p_{g_i}(S) = \hat{A}$ and improved actions $A^{*}$ from the critic \ref{eq:fitness}.
 
\begin{align}
    F_{mse} (g_i, A, S) &= \frac{1}{n} \sum_{j=0}^{n} \left( p_{g_i}(S) - A^*_j \right)^2
    \label{eq:fitness} \\
    F_{var} &= \operatorname{state\_variables}(p_{g_i}) / |S| \\
    \operatorname{fitness} &= (1 - F_{mse}) \times F_{var}
\end{align}

\noindent
with $F_{var}$ a fitness term that encourages programs to look at the state variables (as opposed to producing constants). It is computed by looking at how many state variables the program looks at, divided by the number of state dimensions.

The optimized programs can now be used to predict actions for new states, as the TD3 agent continues. We stress that our method allows the programs to influence exploration (as opposed to \textit{a posteriori} distillation), and that the Genetic Algorithm runs on values that do not require numerous rollouts to be performed in the environment (thus allowing for high sample-efficiency). Our contribution is graphically summarized in Figure \ref{fig:TD3_extended}.

\section{Results}
\subsection{Environment}
\begin{figure}[t]
\centering
\includegraphics[width=0.5\linewidth]{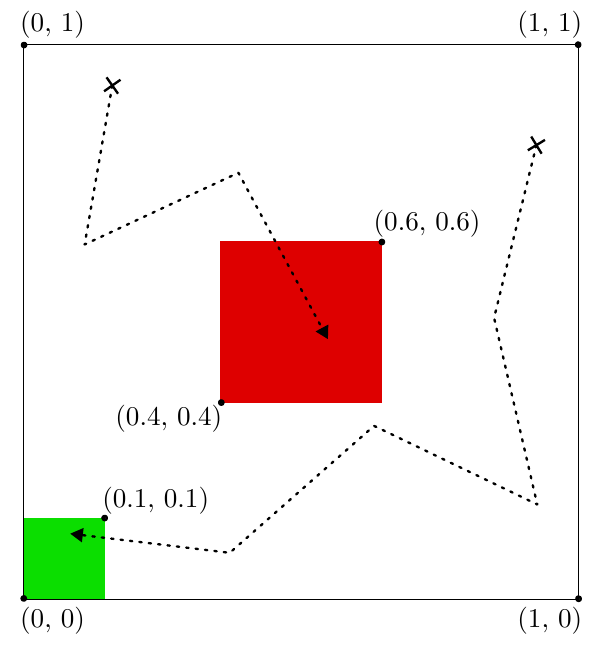}
 \caption{The \texttt{SimpleGoal} environment with the green square the goal region and red the forbidden region. The agent starts at a random position in the environment and tries to reach the goal as quickly as possible.}
\label{fig:simple_goal}
\end{figure}

To validate our approach, we used an environment called \texttt{SimpleGoal} (fig. \ref{fig:simple_goal}). This navigation task is performed in a bounded continuous space of size $1 \times 1$ where the agent needs to take steps towards a goal area located at $x < 0.1, y < 0.1$. The initial starting position is random. The observation space is the current $(x, y)$ coordinate of the agent. The action space is in the range $[-1, 1]$ and defines the change in x and y for the next time-step, with $dx = 0.1 a_0$ and $dy = 0.1 a_1$. At each timestep, the reward $r_t = 10 * (\texttt{old\_distance} - \texttt{new\_distance})$ is calculated based on progress in lowering Euclidean distance towards the goal. If the goal area is reached, an additional reward of 10 is given and the episode terminates. A forbidden area exists at the center of the environment, at coordinates $0.4 < x < 0.6, 0.4 < y < 0.6$. Entering this region terminates the episode with a reward of -10. Otherwise, episodes terminate after 50 time-steps.

\subsection{Training}
To evaluate our method, we trained on several runs using nodes on HPC infrastructure. We got allocated 40 cores out of a a 2x 32-core AMD EPYC 9384X node with 377 GB of memory. A run of 15.000 steps with our method took roughly 8 hours.

% Other methods
To compare to the methods that inspired ours, we set up the same experiments for a vanilla TD3 agent and a pure Genetic Programming approach. The TD3 algorithm is the original one from the CleanRL implementation \cite{huang2022cleanrl}. Our method builds on that TD3 implementation, hence this comparison allows to measure how was TD3 before we introduce our programmatic policy. Learning starts at the same timestep as our method, 2.000. No changes to the original hyperparameters were made.

\begin{figure}
\centering
 \includegraphics[width=\linewidth]{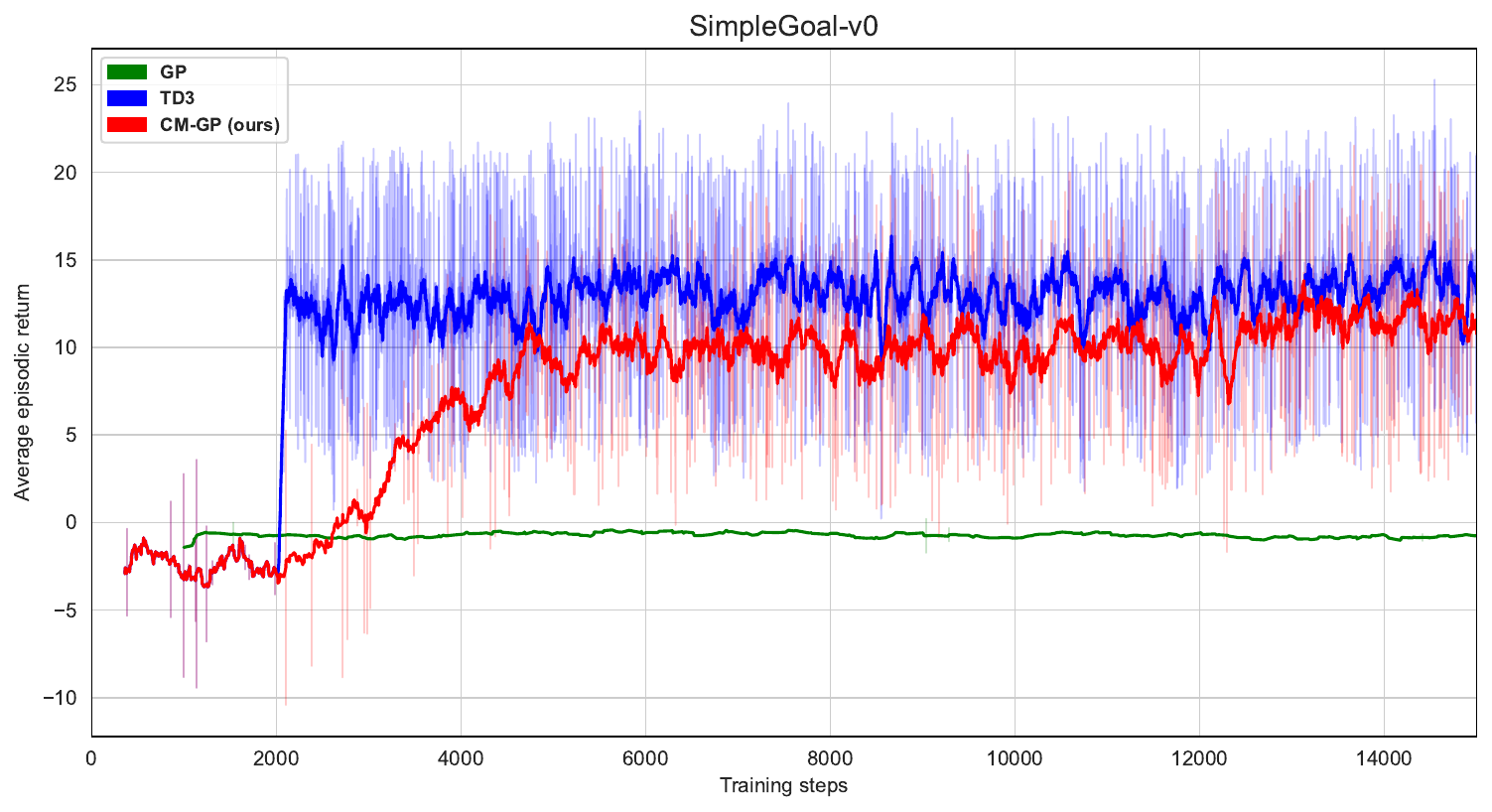}
 \caption{Comparison between vanilla TD3 (blue), Genetic Programming (green) and TD3 + Genetic Programming (ours, red). The intervals indicate the standard error across 5 runs.}
\label{fig:res_comp}
\end{figure}

For the vanilla Genetic Programming (without TD3 critics, using only rollouts), we used the settings as described in table \ref{tab:hparams}. During a learning iteration, the best performing program stands in for the agent in the RL loop. When the policy is updated, the GP process is performed and a new best program is selected. In this case, the fitness function is the performance of one episode in the environment. We already can note that the amount of interactions will increase dramatically with an increase in both \texttt{num\_generations} and \texttt{num\_individuals}. We show our results in figure \ref{fig:res_comp}.

The vanilla GP approach is sample-inefficient, needing an order of magnitude more interactions with the environment to produce good programs. However, when GP is steered by our critic-moderated approach, learning becomes drastically more efficient. This is because of the lack of environment interactions we need to perform to calculate the fitness function since we do this based on the produced gradients of the critic. TD3 has an even higher sample-efficiency. From these results, we can conclude that:

\begin{itemize}
    \item Our method produces explainable programs (see next section) at no cost of final policy quality, and with much higher sample-efficiency than vanilla Genetic Programming.
\end{itemize}

\subsection{Produced programs}
% Plots
We took a selection of produced programs at the end and plotted their behaviour in an arrow plot (fig. \ref{fig:colorspace_comp}). The full programs of each plot are listed in the appendix. At first, we can see that the arrows tend to point towards the goal area in the left bottom corner. The distance of actions taken are quite small, leading to the agent taking a large amount of steps to get to the goal relative to the total distance it has to traverse.
For almost all programs (except \texttt{prog\_1}) the closer the agent is at the goal the larger the step it will take towards it. Since the environment is not strict on going out of range, the agent can take a big step towards the wall and just move along an angle at its edge.
Where \texttt{prog\_2} and \texttt{prog\_4} tend to avoid the pitfall area, \texttt{prog\_1} and \texttt{prog\_3} both have the tendency to enter the area from the right. In the former ones, we also notice a tendency to get stuck in the bottom-right corner.

% Explainability
When we examine the program notations, we can observe some patterns. First, programs tend to be quite complicated in their raw forms, with always-true conditions and a general tendency for producing constants. However, automated or manual constant propagation can be used to make the programs more readable. For all action variables we see the regular incorporation of \texttt{sin} and \texttt{cos}, probably to have a bending curve effect on the action. In general, we notice both action variables produce negative value most of the time, resulting in a direction towards $(0, 0)$.

\begin{figure}[t]
\centering
\includegraphics[width=.24\linewidth]{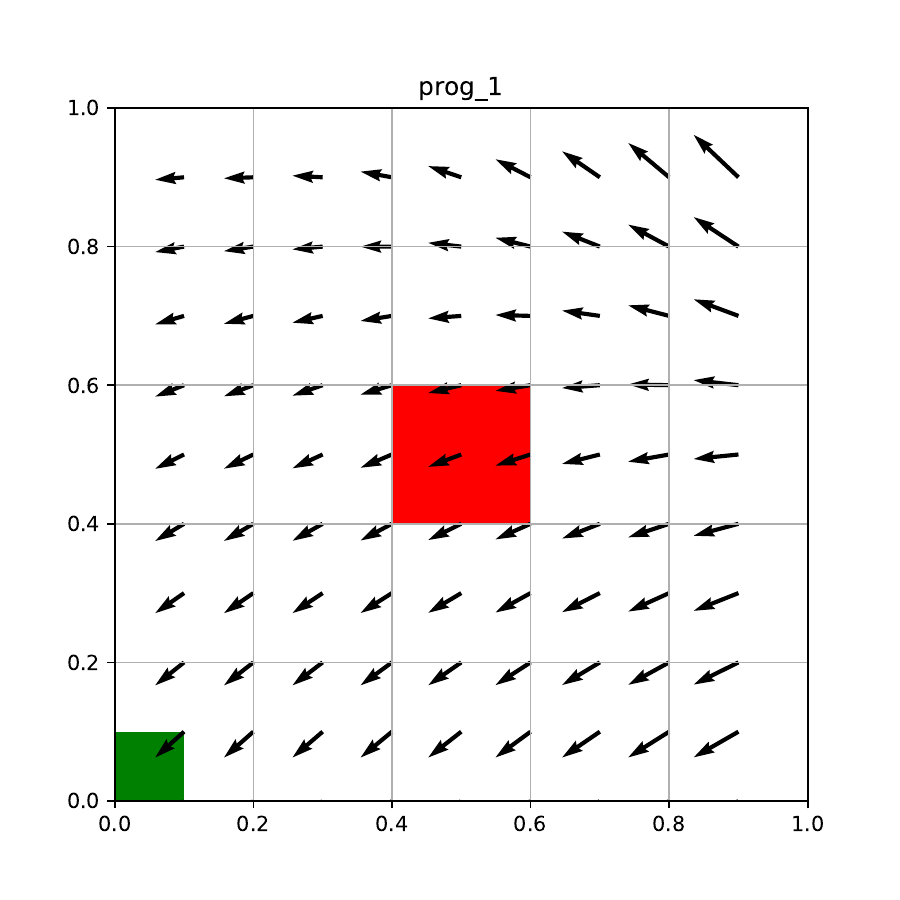}
\includegraphics[width=.24\linewidth]{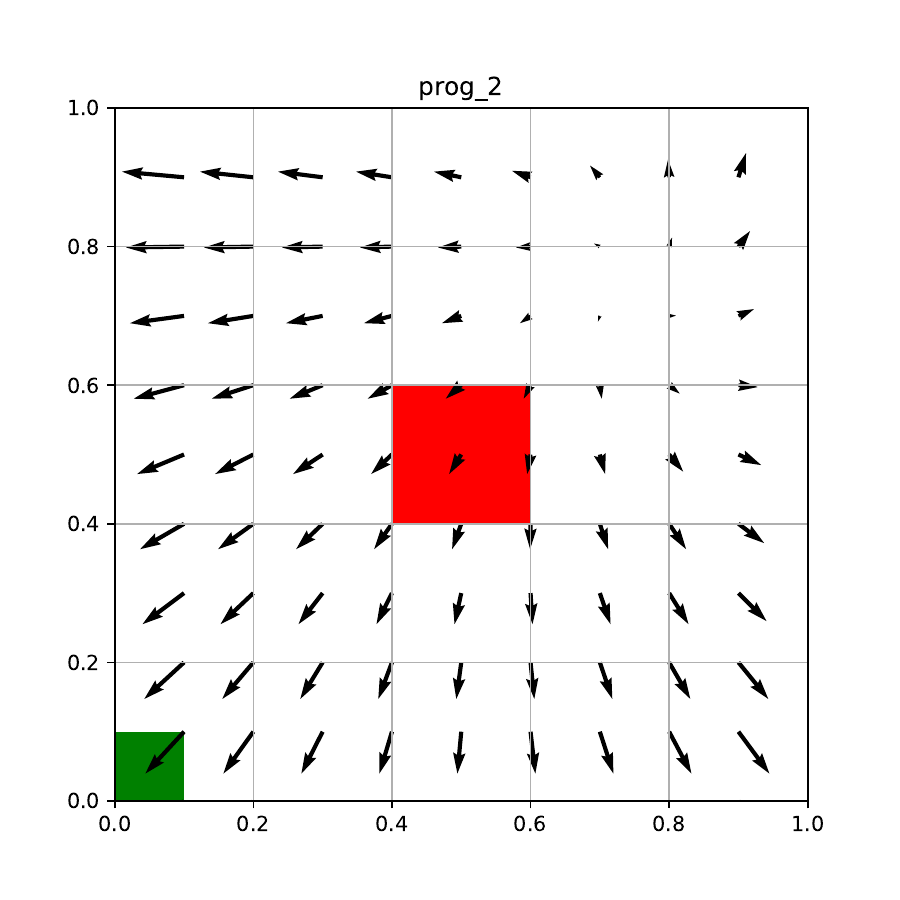}
\includegraphics[width=.24\linewidth]{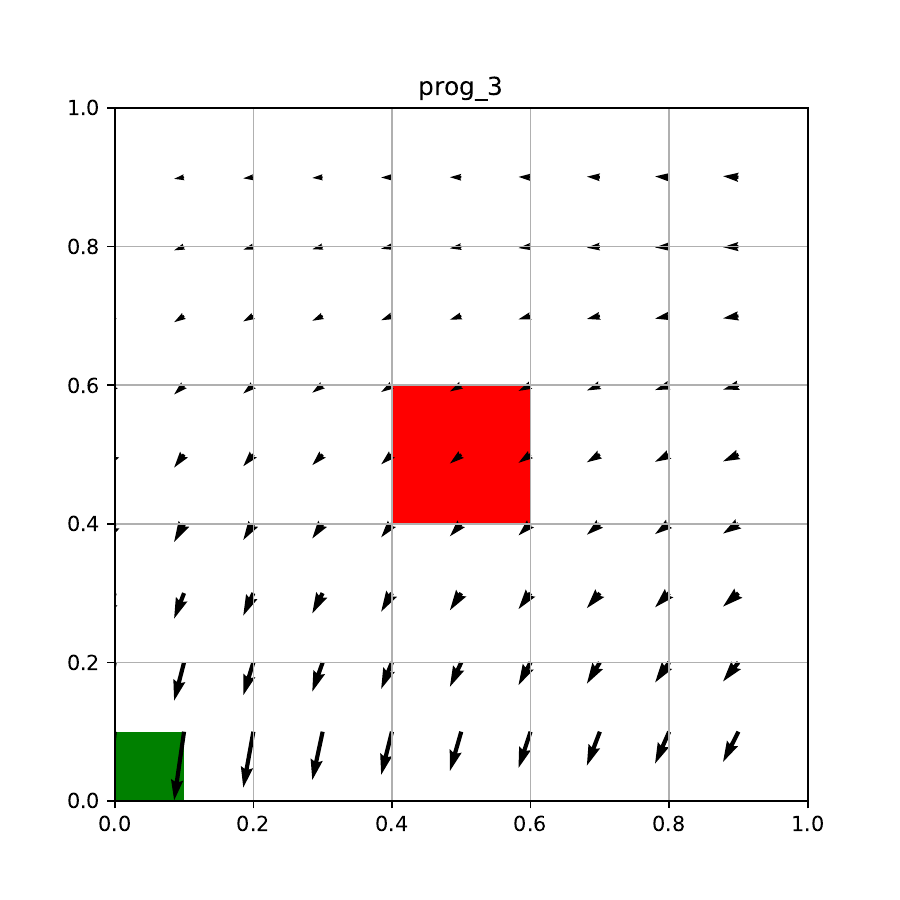}
\includegraphics[width=.24\linewidth]{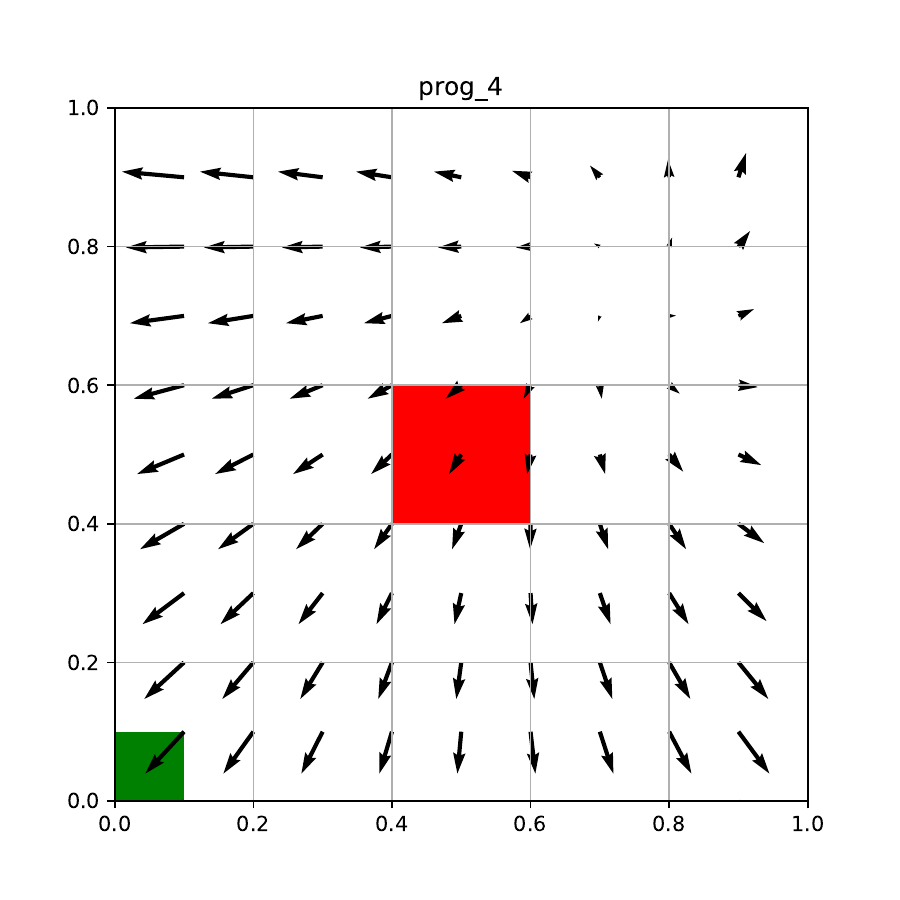}
\caption{Arrow plots indicating a sample of policies on the gridworld. The longer the arrow, the bigger the step taken in the direction it points to. Most programs have a tendency to move towards the goal, most of them avoiding the forbidden area.}
\label{fig:colorspace_comp}
\end{figure}

\section{Conclusion}
We introduced a new method, building on TD3 and Genetic Programming, for generating programs out of Reinforcement Learning agents based on a critic network. The programs are produced and tuned as part of the RL agent learning (they are its policy). Our experiments show that the learned policies are of comparable quality to black-box vanilla TD3 policies, with a sample-efficiency several orders of magnitude higher than Genetic Programming without TD3.

% Improved representation
Our current program representation is quite simple compared to other structures used in GP. Future research avenues include looking into tree-based or graph-based program representations and their dedicated operations. We also designed our algorithm such that the optimization landscape of the GP algorithm is smoothed and without too many local optima. Future program representations may further improve the search landscape, hopefully allowing for more readable yet more expressive policies to be learned in challenging environments.

% Domain specific primitives
Finally, we note that the selection of operators is domain-dependent. Selecting a more suiting set of primitives would benefit the interpretability of the produced programs by the end user.

%Another improvement would be to first find a suiting state-space demarcation based on certain boundaries. For example, states with high value, indicating most actions yield high Q-values, can be separated from regions with lower val

%before growing smaller subprograms in each of these spaces. 

% AST

\begin{credits}
\subsubsection{Acknowledgements} This research received funding from the Flanders Research Foundation via FWO S007723N (CTRLxAI) and FWO G062819N (Explainable Reinforcement Learning). We acknowledge financial support from the Flemish Government (AI Research Program).

\subsubsection{Disclosure of Interests} The authors of this dissemination declare to have no conflict of interest with any other party.

\subsubsection{Reproduction} Code is made available for reproduction at \url{https://github.com/SenneDeproost/CM-GP.git} under MIT licence.

\end{credits}

\vfill
\pagebreak
\printbibliography

\vfill
\pagebreak

\appendix
\section{Experimental setup}

\begin{table}
\centering
\begin{tabular}{|c|c|}
\hline
Hyperparameter                                 & Value\\
\hline
\texttt{num\_genes}                            & 5\\
\texttt{num\_individuals}                      & 50\\
\texttt{num\_generations}                      & 20\\
\texttt{num\_parents\_mating}                  & 20\\
\texttt{mutation\_probability}                 & 0.1\\
\texttt{parent\_selection\_type}               & \texttt{sss}\\
\texttt{crossover\_type}                       & \texttt{single\_point}\\
\texttt{mutation\_type}                        & \texttt{random}\\
\texttt{random\_mutation\_min\_value}          & -10\\
\texttt{random\_mutation\_max\_value}          & 10\\
\texttt{policy\_freq}                          & 128\\

\hline
\end{tabular}
\caption{Used hyperparameters for the evolution strategy}\label{tab:hparams}
\end{table}

\section{Operators}

\begin{table}[H]
\centering
\begin{tabular}{|c|c|c|}
\hline
Operators               &  Description                                                                  & Operands\\
\hline
\texttt{abs / -abs}     &  Absolute value                                                               & 1\\
\texttt{sin / -sin}     &  Sine value                                                                   & 1\\
\texttt{cos / -cos}     &  Cosine value                                                                 & 1\\
\texttt{exp / -exp}     &  Exponent with max operand value 10                                           & 1\\
\texttt{neg}            &  Negation of the value                                                        & 1\\
\texttt{+}              &  Addition                                                                     & 2\\
\texttt{*}              &  Multiplication                                                               & 2\\
\texttt{select}         &  Conditional with test and two cases                                          & 3\\
\texttt{max}            &  Maximum of two operands                                                      & 2\\
\texttt{min}            &  Minimum of two operands                                                      & 2\\
\texttt{id}             &  Identity function                                                            & 1\\
\texttt{reciprocal}     &  Inverse value                                                                & 1\\
\texttt{trunc}          &  Truncation of the value                                                      & 1\\

\hline
\end{tabular}
\caption{Available operators}\label{tab:operators}
\end{table}

\newpage

\section{Produced programs}

The produced programs are listed here in their raw form, and then after constant propagation (given a domain of x in $[-1, 1]$).

\begin{description}
    \item[\texttt{prog\_1}] \hfill \\
        \texttt{a[0] = -exp(max(-sin((x[0] if trunc(x[1]) > 0 else x[1])), x[0]))} \\
        \texttt{a[1] = -cos(((((x[1] + x[0]) + x[1]) * x[0]) * x[1]))}
        \\
    \item[\texttt{prog\_1 simplified}] \hfill \\
        \texttt{a[0] = -exp(x[0])} \\
        \texttt{a[1] = -cos(((((x[1] + x[0]) + x[1]) * x[0]) * x[1]))} \\
        \\
        
    \item[\texttt{prog\_2}] \hfill \\
        \texttt{a[0] = -cos(cos(x[1])) if abs(±66.31885466661134) > 0 else x[0]} \\
        \texttt{a[1] = -abs(cos(max(cos(-sqrt(x[1])), x[0])))} 
        \\
    \item[\texttt{prog\_2 simplified}] \hfill \\
        \texttt{a[0] = -cos(cos(x[1]))} \\
        \texttt{a[1] = -abs(cos(max(cos(-sqrt(x[1])), x[0])))}\\
        \\
        
    \item[\texttt{prog\_3}] \hfill \\
        \texttt{a[0] = -exp(max((max(-abs(x[1]), x[0]) * x[1]), x[0]))} \\
        \texttt{a[1] = -abs(reciprocal(-sqrt(((x[1] + x[0]) * x[1]))))} 
        \\
    \item[\texttt{prog\_3 simplified}] \hfill \\
        \texttt{a[0] = -exp(max((max(-abs(x[1]), x[0]) * x[1]), x[0]))} \\
        \texttt{a[1] = -abs(reciprocal(-sqrt(((x[1] + x[0]) * x[1]))))} \\
        \\
        
    \item[\texttt{prog\_4}] \hfill \\
        \texttt{a[0] = -cos(cos(x[1])) if exp(±64.18861262866074) > 0 else x[0]} \\
        \texttt{a[1] = neg(cos(max(cos(-sqrt(x[1])), x[0])))}
        \\
    \item[\texttt{prog\_4 simplified}] \hfill \\
        \texttt{a[0] = -cos(cos(x[1]))} \\
        \texttt{a[1] = neg(cos(max(cos(-sqrt(x[1])), x[0])))}
        
\end{description}

%\begin{table}[H]
%\centering
%\begin{tabular}{|c|c|c|}
%\hline
%Name  & \texttt{a[0]}                                                               & \texttt{a[1]}\\
%\hline
%\texttt{prog\_1}      & \texttt{-exp(max(-sin((x[0] if trunc(x[1]) > 0 else x[1])), x[0]))}         & \texttt{-cos(((((x[1] + x[0]) + x[1]) * x[0]) * x[1]))}\\
%\hline
%\end{tabular}
%\caption{Available operators}\label{tab:operators}
%\end{table}

\end{document}